\pdfoutput=1

\documentclass[11pt]{article}

\usepackage{coling}

\usepackage{times}
\usepackage{latexsym}

\usepackage[T1]{fontenc}

\usepackage[utf8]{inputenc}

\usepackage{microtype}

\usepackage{inconsolata}

\usepackage{graphicx}

\usepackage{kotex}
\usepackage{xspace}
\usepackage[inline,shortlabels]{enumitem}
\usepackage{amsmath}
\usepackage{tabularray}
\usepackage{booktabs}
\usepackage{tabularx}
\usepackage{multirow, makecell}
\usepackage{amssymb}
\usepackage{todonotes}

\usepackage{booktabs}
\usepackage{graphicx}
\usepackage{pifont}
\definecolor{lp}{HTML}{CBC3E3}
\usepackage{colortbl}
\usepackage{pythonhighlight}
\usepackage{xspace}
\newcommand{\method}{sDPO\xspace}
\usepackage{amsmath}
\usepackage{amsfonts}
\title{sDPO: Don't Use Your Data All at Once}

\newcommand\blfootnote[1]{%
  \begingroup
  \renewcommand\thefootnote{}\footnote{#1}%
  \addtocounter{footnote}{-1}%
  \endgroup
}

\author{Dahyun Kim, Yungi Kim, Wonho Song, Hyeonwoo Kim, Yunsu Kim, Sanghoon Kim\\{\bf \large Chanjun Park$^{\dagger}$}\\
\\
  Upstage AI\\
  \texttt{\normalsize\{kdahyun, eddie, ynot, choco\_9966, yoonsoo, limerobot, chanjun.park\}@upstage.ai}}

\begin{document}
\maketitle
\begin{abstract}
\blfootnote{$^\dagger$ Corresponding Author}
As large language models (LLMs) continue to advance, aligning them with human preferences has become a critical objective. In this paper, we introduce stepwise DPO (\method), an innovative extension of the recently popularized Direct Preference Optimization (DPO) technique for alignment tuning. \method systematically partitions the available preference datasets and applies them incrementally, rather than utilizing the entire dataset simultaneously. This stepwise manner enables the integration of progressively more aligned reference models within the DPO training framework. Our empirical results demonstrate that \method not only enhances the alignment precision of reference models but also significantly improves the overall performance of the final model, surpassing other prominent LLMs with larger parameter counts.
\end{abstract}

\section{Introduction}
Large language models (LLMs) have revolutionized the field of natural language processing (NLP) by undergoing pre-training, supervised fine-tuning, and alignment tuning, with the latter ensuring the safety and usefulness of the model. 
Reinforcement learning (RL) techniques~\cite{christiano2017deep, bai2022constitutional}, such as proximal policy optimization (PPO)~\cite{schulman2017proximal}, are generally used in this alignment phase.

To address the complicated nature of RL in LLM training, direct preference optimization (DPO)~\cite{rafailov2023direct} have been popularized for its simplicity and effectiveness. DPO involves curating preference datasets using human or strong AI (\textit{e.g.,} GPT-4~\cite{openai2023gpt4}) judgement to select chosen and rejected responses from a pool of multiple answers to a given question. Then, the model being trained (\textit{i.e.}, target model) and a separate reference model compute log probabilities of chosen and rejected responses. Finally, the target model is trained by maximizing the difference of the log probability ratios of the target and the reference models for the chosen and rejected answers. However, obtaining these probabilities can be challenging if one wants to use proprietary models like GPT-4 as the reference model, since they do not offer log probabilities for inputs.

\begin{table}[t!]
\centering
\resizebox{1.00\linewidth}{!}{
\begin{tabular}{lcc}
\toprule
Model & Reference Model & H4\\ \midrule
Mistral-7B-OpenOrca & N/A & 65.84 \\
Mistral-7B-OpenOrca + DPO & SFT Base & 68.87\\
Mistral-7B-OpenOrca + DPO & SOLAR-0-70B & 67.86\\
Mistral-7B-OpenOrca + DPO & Intel-7B-DPO & {\bf 70.13}\\ \midrule
OpenHermes-2.5-Mistral-7B & N/A & 66.10 \\
OpenHermes-2.5-Mistral-7B + DPO & SFT Base & 68.41\\
OpenHermes-2.5-Mistral-7B + DPO & SOLAR-0-70B & 68.90\\
OpenHermes-2.5-Mistral-7B + DPO & Intel-7B-DPO & {\bf 69.72}\\
\bottomrule 
\end{tabular}
}
\caption{DPO results in terms of H4 scores for Mistral-7B-OpenOrca and OpenHermes-2.5-Mistral-7B with different reference models. The best results for each SFT base model are shown in bold.}
\label{tab:mistral_openorca}
\end{table}

Thus, in practice, the reference model is simply set as the base SFT model~\cite{tunstall2023zephyr, intel2023orcadpo, ivison2023camels}, which is a much weaker alternative with potentially misaligned preferences.
Through Eq.~\ref{eq:dpo_loss}, we show that the reference model acts as {\it a lower bound} in DPO, \textit{i.e.,} the target model is optimized to be at least as aligned as the reference model.
Thus, we argue that a reference model that is already more aligned will serve as a better lower bound for DPO training, which would be beneficial for the alignment tuning.
One option would be to utilize the plethora of open source models~\cite{tunstall2023zephyr, ivison2023camels} that have already undergone alignment tuning.

\begin{figure*}[t!]
    \centering
    \resizebox{1.0\linewidth}{!}{
\includegraphics{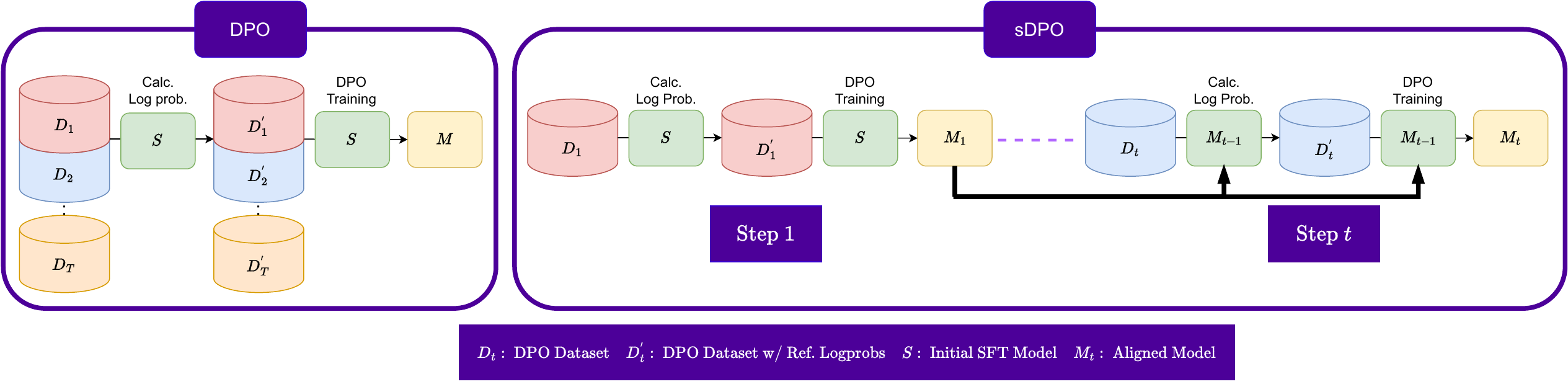}
    }
    \caption{Overview of \method where preference datasets are divided to be used in multiple steps. The aligned model from the previous step is used as the reference and target models for the current step. The reference model is used to calculate the log probabilities and the target model is trained using the preference loss of DPO at each step.}
    \label{fig:method}
\end{figure*}

Note that the above approach may not be feasible due to the absence of such aligned models, or the fact that it renounces control over the reference model, which could lead to safety concerns.
Instead, we propose `stepwise DPO', named \method, where we use the preference datasets (or subsets of a preference dataset) in {\it a step-by-step manner} rather than all at once when undergoing DPO training.
The aligned model in the previous step is used as the reference model for the current step, which results in utilizing a more aligned reference model ({\it i.e.}, a better lower bound).
Empirically, we show that using \method results in a more performant final aligned model as well.

While concurrent works~\citep{yuan2024self} that focus on an iterative pipeline of generating \textit{new} preference data have been proposed, 
our method focuses on utilizing the \textit{currently available} preference datasets.
Thus, our approach is complementary as \method can be easily applied to any preference data and further combination with concurrent works would be an exciting future direction.

\section{Related Work}

\subsection{Large Language Models}
Recent research has highlighted a "scaling law" in the field of context-based language models~\cite{kaplan2020scaling, hernandez2021scaling, anil2023palm}, showing a proportional relationship between the size of the model plus the training data and the resulting performance improvements. Consequently, this has led to the advent of LLMs. In contrast to earlier models, LLMs can perform in-context learning, which includes abilities such as zero-shot learning~\cite{radford2019language} and few-shot learning~\cite{brown2020language}, allowing them to adapt and perform tasks without the need for weight adjustments. These emergent abilities of LLMs, absent in their smaller counterparts, signal a significant evolution in language model capabilities~\cite{wei2022emergent}.

\subsection{Alignment Tuning}
LLMs have been recognized to produce text that may seem linguistically inconsistent to human interpreters because their pretraining is based not on an understanding of human intentions but on a broad spectrum of domain-specific knowledge, as indicated in~\cite{ziegler2019fine}. In an effort to rectify this issue and better mirror human intentions, prior research~\cite{ziegler2019fine} has suggested the adoption of Reinforcement Learning with Human Feedback (RLHF). RLHF seeks to refine the LLM's output by constructing a reward model that aligns with human preferences and applying reinforcement learning to direct the LLM towards selections that garner the most favorable reward metrics. This approach is intended to bolster the safety, decorum, and general excellence of the responses produced by the LLM. Nonetheless, despite showing promising results, RLHF is confronted with challenges, such as the intricate handling of an extensive set of hyperparameters and the necessity to amalgamate several models (policy, value, reward, and reference models).

To address these issues, there have been proposals for supervised fine-tuning methodologies such as RRHF~\cite{yuan2023rrhf}, RAFT~\cite{dong2023raft}, and DPO~\cite{rafailov2023direct}. These methods circumvent the intricacies inherent in reinforcement learning and have been shown to yield empirical results on par with RLHF. Notably, the DPO technique straightforwardly encourages the LLM to favor positive responses and discourage negative ones. DPO has been observed to yield performant learning outcomes, in spite of its uncomplicated training procedure.

Concurrent to our work, \citet{yuan2024self} have developed an iterative framework for generating \textit{new} preference datasets and performing DPO training on the resulting datasets. They empirically demonstrated the superiority of their iterative framework in terms of AlpacaEval 2.0.
In contrast, our work is complementary to the above in the sense that we focus on utilizing the \textit{current} preference data and does not undergo new data generation. 
Thus, our method can also be applied to \citet{yuan2024self} by changing the DPO training part to using sDPO instead. 
Additionally, the evaluation used in \citet{yuan2024self} is also different to ours as we utilize tasks from Open LLM Leaderboard~\cite{open-llm-leaderboard}, EQ Bench~\cite{paech2023eq} and MT Bench~\cite{zheng2023judging} whereas \citet{yuan2024self} uses AlpacaEval 2.0.

\section{Methodology}

\subsection{Preliminary Investigation on Reference Models}
To gauge the importance of using a well-aligned reference model in DPO, we perform preliminary experiments of DPO training with the Ultrafeedback dataset~\cite{cui2023ultrafeedback} on Mistral-7B-OpenOrca~\cite{lian2023mistralorca1} and OpenHermes-2.5-Mistral-7B~\cite{openhermes} as the SFT base model, owing to their excellent performance and small size. We compare the following reference models: i) the SFT base model itself, same as the conventional DPO setup; ii) SOLAR-0-70B~\cite{solar70}, a larger and much more performant model; and iii) Intel-7B-DPO~\cite{intel7b}, an already aligned reference model. The results are summarized in Table~\ref{tab:mistral_openorca}.

As the table shows, using Intel-7B-DPO as the reference model results in the best performance, even better than using SOLAR-0-70B, which is a much larger and performant model.
Thus, whether the reference model is pre-aligned or not plays an important role in the resulting aligned model's performance.
Unfortunately, it is not always possible to use a open sourced pre-aligned model as the reference model due to technical and safety concerns. For instance, such a model may not exist yet or can be susceptible to various domain-specific harmfulness and fairness criteria along with potential data contamination issues.
To circumvent the above, we propose \method, which does not require an external pre-aligned model but uses more aligned reference models, built from the SFT base model, as a part of the training framework.

\subsection{Stepwise DPO}
\label{sec:sdpo}
In \method, we propose to use the available preference datasets in a stepwise manner instead of using them all at once.
Essentially, we partition the preference data into $T$ chunks and perform DPO training $T$ times.
The trained model from the previous step is used as the reference and target models, which means that each of the $T$ DPO training steps function in a similar manner to the conventional DPO setup.
In doing so, we create and utilize intermediary reference models that are more aligned than those thar are used in conventional DPO.
The comparison of the overall flow of DPO and sDPO is presented in Figure~\ref{fig:method}.
\paragraph{Reference model.}
The reference model is used to calculate the log probabilities of the preference dataset.
For each step, only a subset of the total data is used and the reference model is initialized as $M_{t-1}$, \textit{i.e,} the aligned model from the previous step.
The initial reference model is set as $S$, the SFT base model.
This results in using a more aligned reference model than conventional DPO.

\paragraph{Target model.}
For $t > 1$, the target model which is trained using the preference loss of DPO in each step of \method is also initialized as $M_{t-1}$ instead of $S$.
This ensures that the final model trained with \method has been directly trained with the same amount data as a model trained with DPO.

\begin{table*}[t!]
\centering
\resizebox{0.85\linewidth}{!}{
\begin{tabular}{lccccccc}
\toprule
Model & Size & Type  & H4 (Avg.) & ARC & HellaSwag & MMLU & TruthfulQA \\ \midrule
 \cellcolor{lp!60}SOLAR 10.7B + SFT + \method &  \cellcolor{lp!60}$\sim$ 11B&  \cellcolor{lp!60}Alignment-tuned& \cellcolor{lp!60}{\bf 74.31}  & \cellcolor{lp!60}{\bf 71.33} & \cellcolor{lp!60}88.08& \cellcolor{lp!60}65.39& \cellcolor{lp!60}{\bf 72.45} \\
 SOLAR 10.7B + SFT + DPO & $\sim$ 11B & Alignment-tuned & 72.67 & 69.62 & 87.16 & 66.00 & 67.90 \\ \midrule
    Mixtral 8x7B-Instruct-v0.1 & $\sim$ 47B& Alignment-tuned&73.40  &70.22 &87.63&71.16&64.58\\
      SOLAR-0-70B-16bit &  $\sim$ 70B&  Instruction-tuned& 72.93  & 71.08 & 87.89& 70.58&62.25 \\
  Qwen 72B & $\sim$ 72B& Pretrained&72.17  &65.19 &85.94&{\bf 77.37}&60.19\\
      Yi 34B & $\sim$ 34B& Pretrained&70.72  &64.59 &85.69&76.35&56.23\\
 \cellcolor{lp!60}SOLAR 10.7B + SFT&  \cellcolor{lp!60}$\sim$ 11B&  \cellcolor{lp!60}Instruction-tuned& \cellcolor{lp!60} 69.51 & \cellcolor{lp!60}67.32 & \cellcolor{lp!60}85.96& \cellcolor{lp!60}65.95& \cellcolor{lp!60}58.80 \\ 
Mistral 7B-Instruct-v0.2& $\sim$ 7B& Instruction-tuned&69.27  &63.14 &84.88&60.78&68.26\\
         Falcon 180B & $\sim$ 180B& Pretrained&68.57  &69.45 &{\bf 88.86}&70.50&45.47\\
  Mixtral 8x7B-v0.1 & $\sim$ 47B& Pretrained&67.78  &66.04 &86.49&71.82&46.78\\
  Llama 2 70B & $\sim$ 70B& Pretrained&67.35  &67.32 &87.33&69.83&44.92\\
Zephyr & $\sim$ 7B& Alignment-tuned & 66.36 & 62.03 & 84.52 & 61.44 & 57.44 \\
Qwen 14B & $\sim$ 14B& Pretrained&64.85  &58.28 &83.99&67.70&49.43\\
 \cellcolor{lp!60} SOLAR 10.7B &  \cellcolor{lp!60}$\sim$ 11B&  \cellcolor{lp!60}Pretrained& \cellcolor{lp!60}64.27  & \cellcolor{lp!60}61.95 & \cellcolor{lp!60}84.60& \cellcolor{lp!60}65.48& \cellcolor{lp!60}45.04 \\

Mistral 7B& $\sim$ 7B& Pretrained&62.40  &59.98 &83.31&64.16&42.15\\
\bottomrule 
\end{tabular}
}
\caption{Performance comparison of applying \method or DPO to SOLAR 10.7B + SFT against various top performing models. Size is shown in units of billions of parameters and type is reported as one of \{`Pretrained', `Instruction-tuned', `Alignment-tuned'\}. Models based on SOLAR 10.7B are shown in purple color. The best scores in each column are shown in bold.}
\label{tab:main_result}
\end{table*}

\paragraph{Mathematical explanation.}
To gain a deeper understanding of sDPO, we rearrange the DPO loss from ~\cite{rafailov2023direct}, as follows:
\begin{equation}
\resizebox{0.85\linewidth}{!}{%
\begin{math}
\begin{aligned}
     & \mathcal{L}_{\text{DPO}}(\pi_\theta, \pi_{ref}) \\
     & = - \mathbb{E}_{(x, y_w, y_l)\sim \mathcal{D}}\left[\log{\sigma\left(\beta\log{{\pi_\theta(y_w|x) \over \pi_{ref}(y_w|x)}} - \beta \log{{\pi_\theta(y_l|x) \over \pi_{ref}(y_l|x)}}\right)}\right] \\
     & = - \mathbb{E}_{(x, y_w, y_l)\sim \mathcal{D}}\left[\log{\sigma\left(\beta\cdot (\gamma_{\pi_\theta}(x, y_w, y_l) - \gamma_{\pi_{ref}}(x, y_w, y_l)\right)}\right],
\end{aligned}
\end{math}
}
\label{eq:dpo_loss}
\end{equation}
where $D$ is the preference dataset, $x$ is the question, $y_w$ and $y_l$ are the chosen and rejected answers respectively, $\theta$ is the learnable parameters of the model, and $\gamma_\pi(x, y_w, y_l) = \log{{\pi(y_w|x) \over \pi(y_l|x)}}$, \textit{i.e.,} the logratio of the chosen and rejected samples w.r.t. the policy $\pi$.
As $\log\sigma(\cdot)$ is a monotonically increasing function and $\gamma_{\pi_{ref}}$ is fixed before training, the minimization of $\mathcal{L}_{\text{DPO}}(\pi_\theta, \pi_{ref})$ leads to $\gamma_{\pi_{\theta}} > \gamma_{\pi_{ref}}$ on average.
Thus, $\gamma_{\pi_{ref}}$ can be understood as a lower bound defined by the reference model, of which the target model is trained such that $\gamma_{\pi_{\theta}} > \gamma_{\pi_{ref}}$.
In sDPO, $\gamma_{\pi_{ref}}$ increases as the steps progress because the reference model that defines it is more and more aligned.
Hence, $\gamma_{\pi_{ref}}$ becomes a stricter lower bound as the steps pass, inducing a {\it curriculum learning} from easy to hard optimization tasks.
Thus, the target model is being trained to learn stricter preferences as the steps progress in \method.

\paragraph{Data partitioning strategy.}
The method for partitioning the preference data into $T$ chunks is also important in \method.
One option would be to pool all the data from different dataset sources and perform random sampling.
However, we argue that partitioning the data such that earlier chunks are comprised of easier preference data would be more aligned with inducing a curriculum learning of easy to hard optimization in \method.

To that end, we propose to use easy-to-hard data partitioning by the following method.
Using $M_0$, the initial target model, we calculate the reward accuracy, \textit{i.e.,} the percentage of samples in which the target model scores higher rewards for preferred samples, for the different dataset sources.
The dataset sources are sorted in descending order of the reward accuracy, which are then used as the $T$ chunks in \method.
Thus, if we have $N$ dataset sources, we would have a total of $N$ chunks, where earlier chunks would contain easier samples as measured by the reward accuracy.

\section{Experiments}
\subsection{Experimental Setup}
\paragraph{Training details.} We use a supervised fine-tuned SOLAR 10.7B~\cite{kim2023solar} as our SFT base model $S$ as it delivers excellent performance with its uncommon yet relatively small 10.7B size.
Note that we do not need a separate reference model as it is initialized as $M_{t-1}$, the final trained model from the previous step. 
We use OpenOrca~\cite{mukherjee2023orca} ($\sim12K$ samples) and Ultrafeedback Cleaned ($\sim60K$ samples)~\cite{cui2023ultrafeedback, ivison2023camels} as our preference datasets.
The training hyper-parameters follow that of~\citet{tunstall2023zephyr}.
Using the easy-to-hard partitioning, we use OpenOrca as dataset $D_1$ and Ultrafeedback Cleaned as dataset $D_2$.

\paragraph{Evaluation.} We mainly utilize four log-probability tasks in the HuggingFace Open LLM Leaderboard~\cite{open-llm-leaderboard}: ARC~\cite{clark2018think}, HellaSWAG~\cite{zellers2019hellaswag}, MMLU~\cite{hendrycks2020measuring}, TruthfulQA~\cite{lin2022truthfulqa}.
We also report the average scores for the four tasks, which is denoted as H4.
Note that these tasks do not require the model to actually generate a new answer to the question.
Rather, the log-probability of a pre-defined answer is measured instead.

To augment the above potential downside of log-probability benchmarks, we also incorporate generation benchmarks such as EQ Bench~\cite{paech2023eq} and MT Bench~\cite{zheng2023judging}, where a model is prompted to generate an answer to a question.
As such, MT Bench and EQ Bench both strongly correlate with the Chatbot Arena ELO~\cite{zheng2023judging,chiang2024chatbotarenaopenplatform}, one of the most widely recognized open-world LLM evaluation system.

\subsection{Main Results}
Evaluation results for applying \method to the SFT base model, along with results for other top-performing models are shown in Table~\ref{tab:main_result}.
Applying \method on SOLAR 10.7B + SFT further increases the H4 score to $74.31$, an improvement of $+4.80$.
Notably, `SOLAR 10.7B + SFT + \method' outperforms other larger models such as Mixtral 8x7B-Instruct-v0.1, despite the smaller number of parameters.
This highlights that effective alignment tuning could be the key to unlocking next level performance for smaller LLMs.
Further, applying \method results in substantially higher score of $72.45$ for TruthfulQA, which demonstrates the effectiveness of the alignment tuning process. 
We also present additional results in Table~\ref{tab:additional_main_result} of Section~\ref{sec:additional_results} on the EQ Bench~\cite{paech2023eq}, which is a generation task with high correlation with the Chatbot Arena ELO~\cite{zheng2023judging}.
The additional results indicate the superiority of \method over DPO in improving generation task performance as well.

\begin{figure}[t!]
    \centering
    \resizebox{1.0\linewidth}{!}{
    \includegraphics{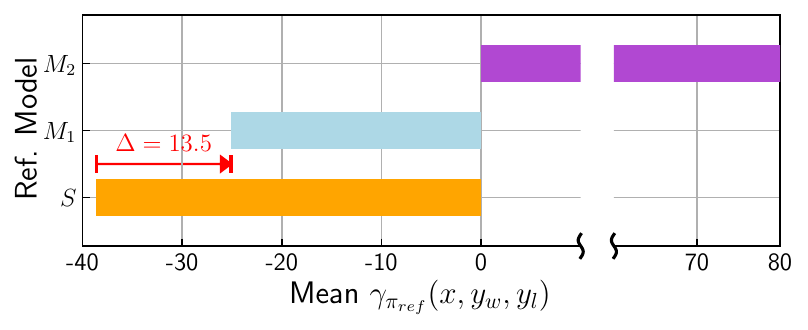}
    }
    \caption{Mean $\gamma_{\pi_{ref}}$ on Ultrafeedback Cleaned dataset for different reference models $S,M_1,$ and $M_2$. Note that the x-axis is in log scale.}
    \label{fig:logratio}
\end{figure}
\subsection{Ablation Studies Against DPO}
We also report evaluation results for ablating \method with traditional DPO in Table~\ref{tab:main_result}.
`SOLAR 10.7B + SFT + DPO' uses all the DPO data at once, \textit{i.e.,} $D_1 + D_2$, same as the conventional DPO training setup.

We can see that using \method over DPO results in a higher H4 score overall, with noticeable improvements in ARC and TruthfulQA scores.
Therefore, we believe \method could function as a drop-in replacement for DPO training with better performance.

\subsection{Reference Models in \method}

\begin{figure}[t!]
    \centering
    \resizebox{1.0\linewidth}{!}{
    \includegraphics{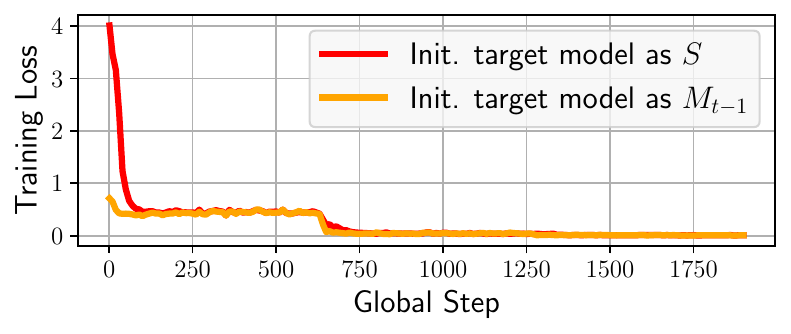}
    }
    \caption{Loss curve comparison in step 2 of \method for different initializations of the target model.}
    \label{fig:init}
\end{figure}

\paragraph{Effectiveness of \method in terms of alignment tuning.}
In Sec.~\ref{sec:sdpo}, we explain that the reference models in \method are more aligned, resulting in higher $\gamma_{\pi_{ref}}$, \textit{i.e.,} a stricter lower bound.
We verify the above empirically in Figure~\ref{fig:logratio} by comparing the mean $\gamma_{\pi_{ref}}$ on the Ultrafeedback Cleaned dataset for the reference models in steps 1 and 2 of \method, \textit{i.e.,} $S$ and $M_1$.
Note that these two models have not been trained on the aforementioned dataset.
Using the SFT base model $S$ as the reference model, the mean of $\gamma_{\pi_{ref}}$ is $-38.60$.
On the other hand, using the aligned model $M_1$ from step 1 of \method as the reference model, the mean of $\gamma_{\pi_{ref}}$ is $-25.10$, an increase of $13.50$ in \textit{log scale}.
Thus, a single step of \method greatly increases $\gamma_{\pi_{ref}}$, which results in a more performant aligned model as seen in Table~\ref{tab:main_result}.

\paragraph{Adopting open source models as reference models could be dangerous.}
We also show mean $\gamma_{\pi_{ref}}$ of $M_2$, the aligned model from step 2 of \method.
Unlike $S$ and $M_1$, $M_2$ is trained on the Ultrafeedback Cleaned dataset, \textit{i.e.,} $M_2$ is used as a reference model on data that was {\it already used to train it}.
Note that such a case could happen commonly when adopting various open source models as reference models.
This is because the datasets that were used in training those models are often unclear and could overlap with the preference datasets unintentionally.
Mean $\gamma_{\pi_{ref}}$ of $M_2$ is $84.35$, which is staggeringly higher than either $S$ or $M_1$. 
The strikingly high value for $M_2$ likely points to overfitting of $M_2$ to the Ultrafeedback Cleaned dataset.
Note that utilizing such an absurdly high value of $\gamma_{\pi_{ref}}$ as the lower bound in DPO training may be undesirable.
This result highlights the potential danger of merely adopting open source models as reference models instead of using \method.

\begin{table*}[t!]
\centering
\resizebox{0.8\linewidth}{!}{
\begin{tabular}{lccccc}
\toprule
Model & H4 (Avg.) & ARC & HellaSwag & MMLU & TruthfulQA \\ \midrule
 \cellcolor{lp!60} SOLAR 10.7B + SFT + \method & \cellcolor{lp!60}{\bf 74.31}  & \cellcolor{lp!60}{\bf 71.33} & \cellcolor{lp!60}{\bf88.08}& \cellcolor{lp!60}65.39& \cellcolor{lp!60}{\bf 72.45} \\
 SOLAR 10.7B + SFT + \method Rand. & 72.56 & 69.20 & 87.27 & {\bf65.96} & 67.81 \\
\bottomrule 
\end{tabular}
}
\caption{Comparison between the easy-to-hard and random partitioning strategies. `SOLAR 10.7B + SFT + \method' uses the easy-to-hard partitioning whereas `SOLAR 10.7B + SFT + \method Rand.' denotes \method with random partitioning instead. Easy-to-hard partitioning is better than random partitioning. The best scores are shown in bold.}
\label{tab:easy2hard}
\end{table*}
\subsection{Target Model Initialization in \method}
One option for target model initialization in \method is to use $S$, the initial SFT base model, for {\it all steps}.
However, such initialization results in the final model trained with \method seeing less data than using DPO instead.
Further, the target model and the reference model become more and more different as the steps progress, which is a deviation from the original DPO setup and risks losing the theoretical benefits of DPO.

To concretely investigate such potential issues, we visualize the loss curves for initializing the target model as $S$ in Figure~\ref{fig:init}.
We observe that the initial loss value is much higher when compared to initializing the target model as $M_{t-1}$, \textit{i.e.,} the same as the reference model and adhering to the DPO convention.
As using $M_{t-1}$ the target model means that each {\it step} of \method is using the same setup as DPO, the loss curves are much more stable and desirable.
Thus, for stable training, initializing the target model as $M_{t-1}$ was chosen for \method.



\subsection{Easy-to-Hard Data Partitioning}

The effectiveness of the easy-to-hard data partitioning used in \method is demonstrated in Table~\ref{tab:easy2hard}.
Note that we use OpenOrca as $D_1$ and Ultrafeedback Cleaned as $D_2$.
As `SOLAR 10.7B + SFT + \method', which uses the easy-to-hard partitioning, is more performant than`SOLAR 10.7B + SFT + \method Rand.', which uses random partitioning, the proposed easy-to-hard data partitioning is more effective for \method.

\subsection{Additional Results on Generation Tasks}
\label{sec:additional_results}

\begin{table}[t!]
\centering
\resizebox{1.0\linewidth}{!}{
\begin{tabular}{lccc}
\toprule
Model & EQ Bench & MT Bench\\ \midrule
 \cellcolor{lp!60}SOLAR 10.7B + SFT + \method & \cellcolor{lp!60}{\bf 68.83} &\cellcolor{lp!60}{\bf  7.43}\\
 SOLAR 10.7B + SFT + DPO & 61.02 & 7.35\\ \midrule
 SOLAR 10.7B + SFT&  60.48 & 7.14\\
\bottomrule 
\end{tabular}
}
\caption{Additional results on EQ Bench~\cite{paech2023eq} and MT Bench~\cite{zheng2023judging}, both of which are generation tasks that highly correlate with the Chatbot Arena ELO~\cite{zheng2023judging,chiang2024chatbotarenaopenplatform}. The best scores for both benchmarks are shown in bold.}
\label{tab:additional_main_result}
\end{table}

In Table~\ref{tab:additional_main_result}, we also report results for EQ Bench~\cite{paech2023eq} and MT Bench~\cite{zheng2023judging} for the SFT base model and the models obtained by applying DPO and \method on the SFT base model.

For EQ Bench, we use the version without the revision prompt.
We note that the EQ Bench requires the models to generate an answer that can be parsed with a pre-defined template for evaluation, which could be said to measure distinct capabilities of LLMs from the log-probability benchmarks shown in Table~\ref{tab:main_result}.
While applying DPO only mildly improves the performance from the SFT base model, applying \method improves the performance significantly by over $+8\%$, indicating the effectivenss in which \method improves the generation capabilities compared to DPO.

As for MT Bench, we note that using \method achieves the best score of 7.43 amongst the compared models.
Notably, applying \method to the SFT base model improves the MT Bench score by a non-trivial margin of $+0.29$.
Applying DPO to the SFT base model also improves the MT Bench score, but not by more than that of applying \method.

\section{Conclusion}

We propose \method, an extension of DPO for aligning LLMs. Unlike traditional DPO, \method employs a stepwise approach, using subsets of preference data sequentially. This method leverages the aligned model from the previous step as the reference for the current step, ensuring progressively better alignment. Our experiments demonstrate that \method significantly outperforms conventional DPO in terms of both log-probability benchmarks such as ARC, HellaSWAG, MMLU, and TruthfulQA, as well as generation benchmarks such as EQ Bench and MT Bench. Additionally, \method enhances model alignment, as indicated by higher mean $\gamma_{\pi_{ref}}$ values, showing improved alignment with human preferences. The stepwise nature of \method simplifies the training process and aligns with curriculum learning principles, facilitating a structured optimization path. By using existing preference datasets more effectively, \method results in higher performance and better-aligned language models. This approach has the potential to transform alignment tuning, offering a robust framework for future research in LLMs.

\section*{Limitations}
While we have demonstrated the effectiveness of employing easy-to-hard data partitioning of different datasets in distinct stages of \method, identifying a more performant strategy for segmenting more intricate preference data collections remains an area for further exploration.

Furthermore, our experiments predominantly utilized SOLAR 10.7B models, driven by the state-of-the-art performance at the time of experimentation along with its relatively 10.7 billion parameter size.
Although as SOLAR 10.7B models are also based on the Llama-2 architecture with our results likely to transfer to other similar decoder only transformer models, more experiments using other models would be beneficial.

Additionally, as with most research on LLMs, we operated within our limitations in computational resources. Although this focus has yielded significant insights, expanding our experimental framework to incorporate a broader range of Large Language Models (LLMs) could potentially unveil more comprehensive understanding of the strengths and limitations of \method. Such an expansion would allow for a more robust comparison across different model architectures and sizes, further enriching our findings.

Evaluating the efficacy of LLMs is an evolving challenge in the field. In our study, we primarily employed tasks from the Huggingface Open LLM Leaderboard as benchmarks for evaluation along with EQ Bench and MT Bench. While this provided comparative results, future research could benefit from incorporating a wider array of tasks and benchmarks. These could include tasks that judge actual human or strong AI preference alignment. Such additional evaluation would not only enhance the validity of our findings but also contribute to the broader discourse on LLM assessment methodologies.

\section*{Ethics Statement}
In this study, we strictly adhered to ethical standards in the conduct of our research. Our experiments were based entirely on open models and open datasets, ensuring transparency and accessibility. We took meticulous care to avoid any biases or data contamination, thereby maintaining the integrity of our research process. The experimental environment was rigorously designed to be objective, ensuring that all comparisons conducted were fair and impartial. This approach reinforces the reliability and validity of our findings, contributing positively to the field while upholding the highest ethical standards. We confirmed that all the data used in our experiments were free of licensing issues.

\bibliography{anthology, custom}

\begin{thebibliography}{33}
\providecommand{\natexlab}[1]{#1}

\bibitem[{Anil et~al.(2023)Anil, Dai, Firat, Johnson, Lepikhin, Passos, Shakeri, Taropa, Bailey, Chen et~al.}]{anil2023palm}
Rohan Anil, Andrew~M Dai, Orhan Firat, Melvin Johnson, Dmitry Lepikhin, Alexandre Passos, Siamak Shakeri, Emanuel Taropa, Paige Bailey, Zhifeng Chen, et~al. 2023.
\newblock Palm 2 technical report.
\newblock \emph{arXiv preprint arXiv:2305.10403}.

\bibitem[{Bai et~al.(2022)Bai, Kadavath, Kundu, Askell, Kernion, Jones, Chen, Goldie, Mirhoseini, McKinnon et~al.}]{bai2022constitutional}
Yuntao Bai, Saurav Kadavath, Sandipan Kundu, Amanda Askell, Jackson Kernion, Andy Jones, Anna Chen, Anna Goldie, Azalia Mirhoseini, Cameron McKinnon, et~al. 2022.
\newblock Constitutional ai: Harmlessness from ai feedback.
\newblock \emph{arXiv preprint arXiv:2212.08073}.

\bibitem[{Beeching et~al.(2023)Beeching, Fourrier, Habib, Han, Lambert, Rajani, Sanseviero, Tunstall, and Wolf}]{open-llm-leaderboard}
Edward Beeching, Clémentine Fourrier, Nathan Habib, Sheon Han, Nathan Lambert, Nazneen Rajani, Omar Sanseviero, Lewis Tunstall, and Thomas Wolf. 2023.
\newblock Open llm leaderboard.
\newblock \url{https://huggingface.co/spaces/HuggingFaceH4/open_llm_leaderboard}.

\bibitem[{Brown et~al.(2020)Brown, Mann, Ryder, Subbiah, Kaplan, Dhariwal, Neelakantan, Shyam, Sastry, Askell et~al.}]{brown2020language}
Tom Brown, Benjamin Mann, Nick Ryder, Melanie Subbiah, Jared~D Kaplan, Prafulla Dhariwal, Arvind Neelakantan, Pranav Shyam, Girish Sastry, Amanda Askell, et~al. 2020.
\newblock Language models are few-shot learners.
\newblock \emph{Advances in neural information processing systems}, 33:1877--1901.

\bibitem[{Chiang et~al.(2024)Chiang, Zheng, Sheng, Angelopoulos, Li, Li, Zhang, Zhu, Jordan, Gonzalez, and Stoica}]{chiang2024chatbotarenaopenplatform}
Wei-Lin Chiang, Lianmin Zheng, Ying Sheng, Anastasios~Nikolas Angelopoulos, Tianle Li, Dacheng Li, Hao Zhang, Banghua Zhu, Michael Jordan, Joseph~E. Gonzalez, and Ion Stoica. 2024.
\newblock \href {https://arxiv.org/abs/2403.04132} {Chatbot arena: An open platform for evaluating llms by human preference}.
\newblock \emph{Preprint}, arXiv:2403.04132.

\bibitem[{Christiano et~al.(2017)Christiano, Leike, Brown, Martic, Legg, and Amodei}]{christiano2017deep}
Paul~F Christiano, Jan Leike, Tom Brown, Miljan Martic, Shane Legg, and Dario Amodei. 2017.
\newblock Deep reinforcement learning from human preferences.
\newblock \emph{Advances in neural information processing systems}, 30.

\bibitem[{Clark et~al.(2018)Clark, Cowhey, Etzioni, Khot, Sabharwal, Schoenick, and Tafjord}]{clark2018think}
Peter Clark, Isaac Cowhey, Oren Etzioni, Tushar Khot, Ashish Sabharwal, Carissa Schoenick, and Oyvind Tafjord. 2018.
\newblock Think you have solved question answering? try arc, the ai2 reasoning challenge.
\newblock \emph{arXiv preprint arXiv:1803.05457}.

\bibitem[{Cui et~al.(2023)Cui, Yuan, Ding, Yao, Zhu, Ni, Xie, Liu, and Sun}]{cui2023ultrafeedback}
Ganqu Cui, Lifan Yuan, Ning Ding, Guanming Yao, Wei Zhu, Yuan Ni, Guotong Xie, Zhiyuan Liu, and Maosong Sun. 2023.
\newblock Ultrafeedback: Boosting language models with high-quality feedback.
\newblock \emph{arXiv preprint arXiv:2310.01377}.

\bibitem[{Dong et~al.(2023)Dong, Xiong, Goyal, Pan, Diao, Zhang, Shum, and Zhang}]{dong2023raft}
Hanze Dong, Wei Xiong, Deepanshu Goyal, Rui Pan, Shizhe Diao, Jipeng Zhang, Kashun Shum, and Tong Zhang. 2023.
\newblock Raft: Reward ranked finetuning for generative foundation model alignment.
\newblock \emph{arXiv preprint arXiv:2304.06767}.

\bibitem[{Hendrycks et~al.(2020)Hendrycks, Burns, Basart, Zou, Mazeika, Song, and Steinhardt}]{hendrycks2020measuring}
Dan Hendrycks, Collin Burns, Steven Basart, Andy Zou, Mantas Mazeika, Dawn Song, and Jacob Steinhardt. 2020.
\newblock Measuring massive multitask language understanding.
\newblock In \emph{International Conference on Learning Representations}.

\bibitem[{Hernandez et~al.(2021)Hernandez, Kaplan, Henighan, and McCandlish}]{hernandez2021scaling}
Danny Hernandez, Jared Kaplan, Tom Henighan, and Sam McCandlish. 2021.
\newblock Scaling laws for transfer.
\newblock \emph{arXiv preprint arXiv:2102.01293}.

\bibitem[{Intel(2023{\natexlab{a}})}]{intel7b}
Intel. 2023{\natexlab{a}}.
\newblock Intel/neural-chat-7b-v3-1.
\newblock \url{https://huggingface.co/Intel/neural-chat-7b-v3-1}.

\bibitem[{Intel(2023{\natexlab{b}})}]{intel2023orcadpo}
Intel. 2023{\natexlab{b}}.
\newblock \href {https://medium.com/intel-analytics-software/the-practice-of-supervised-finetuning-and-direct-preference-optimization-on-habana-gaudi2-a1197d8a3cd3} {Supervised fine-tuning and direct preference optimization on intel gaudi2}.

\bibitem[{Ivison et~al.(2023)Ivison, Wang, Pyatkin, Lambert, Peters, Dasigi, Jang, Wadden, Smith, Beltagy, and Hajishirzi}]{ivison2023camels}
Hamish Ivison, Yizhong Wang, Valentina Pyatkin, Nathan Lambert, Matthew Peters, Pradeep Dasigi, Joel Jang, David Wadden, Noah~A. Smith, Iz~Beltagy, and Hannaneh Hajishirzi. 2023.
\newblock \href {https://arxiv.org/abs/2311.10702} {Camels in a changing climate: Enhancing lm adaptation with tulu 2}.
\newblock \emph{Preprint}, arXiv:2311.10702.

\bibitem[{Kaplan et~al.(2020)Kaplan, McCandlish, Henighan, Brown, Chess, Child, Gray, Radford, Wu, and Amodei}]{kaplan2020scaling}
Jared Kaplan, Sam McCandlish, Tom Henighan, Tom~B Brown, Benjamin Chess, Rewon Child, Scott Gray, Alec Radford, Jeffrey Wu, and Dario Amodei. 2020.
\newblock Scaling laws for neural language models.
\newblock \emph{arXiv preprint arXiv:2001.08361}.

\bibitem[{Kim et~al.(2023)Kim, Park, Kim, Lee, Song, Kim, Kim, Kim, Lee, Kim, Ahn, Yang, Lee, Park, Gim, Cha, Lee, and Kim}]{kim2023solar}
Dahyun Kim, Chanjun Park, Sanghoon Kim, Wonsung Lee, Wonho Song, Yunsu Kim, Hyeonwoo Kim, Yungi Kim, Hyeonju Lee, Jihoo Kim, Changbae Ahn, Seonghoon Yang, Sukyung Lee, Hyunbyung Park, Gyoungjin Gim, Mikyoung Cha, Hwalsuk Lee, and Sunghun Kim. 2023.
\newblock \href {https://arxiv.org/abs/2312.15166} {Solar 10.7b: Scaling large language models with simple yet effective depth up-scaling}.
\newblock \emph{Preprint}, arXiv:2312.15166.

\bibitem[{Lian et~al.(2023)Lian, Goodson, Wang, Pentland, Cook, Vong, and "Teknium"}]{lian2023mistralorca1}
Wing Lian, Bleys Goodson, Guan Wang, Eugene Pentland, Austin Cook, Chanvichet Vong, and "Teknium". 2023.
\newblock Mistralorca: Mistral-7b model instruct-tuned on filtered openorcav1 gpt-4 dataset.
\newblock \url{https://huggingface.co/Open-Orca/Mistral-7B-OpenOrca}.

\bibitem[{Lin et~al.(2022)Lin, Hilton, and Evans}]{lin2022truthfulqa}
Stephanie Lin, Jacob Hilton, and Owain Evans. 2022.
\newblock Truthfulqa: Measuring how models mimic human falsehoods.
\newblock In \emph{Proceedings of the 60th Annual Meeting of the Association for Computational Linguistics (Volume 1: Long Papers)}, pages 3214--3252.

\bibitem[{Mukherjee et~al.(2023)Mukherjee, Mitra, Jawahar, Agarwal, Palangi, and Awadallah}]{mukherjee2023orca}
Subhabrata Mukherjee, Arindam Mitra, Ganesh Jawahar, Sahaj Agarwal, Hamid Palangi, and Ahmed Awadallah. 2023.
\newblock Orca: Progressive learning from complex explanation traces of gpt-4.
\newblock \emph{arXiv preprint arXiv:2306.02707}.

\bibitem[{OpenAI(2023)}]{openai2023gpt4}
OpenAI. 2023.
\newblock \href {https://arxiv.org/abs/2303.08774} {Gpt-4 technical report}.
\newblock \emph{Preprint}, arXiv:2303.08774.

\bibitem[{Paech(2023)}]{paech2023eq}
Samuel~J Paech. 2023.
\newblock Eq-bench: An emotional intelligence benchmark for large language models.
\newblock \emph{arXiv preprint arXiv:2312.06281}.

\bibitem[{Radford et~al.(2019)Radford, Wu, Child, Luan, Amodei, Sutskever et~al.}]{radford2019language}
Alec Radford, Jeffrey Wu, Rewon Child, David Luan, Dario Amodei, Ilya Sutskever, et~al. 2019.
\newblock Language models are unsupervised multitask learners.
\newblock \emph{OpenAI blog}, 1(8):9.

\bibitem[{Rafailov et~al.(2023)Rafailov, Sharma, Mitchell, Ermon, Manning, and Finn}]{rafailov2023direct}
Rafael Rafailov, Archit Sharma, Eric Mitchell, Stefano Ermon, Christopher~D Manning, and Chelsea Finn. 2023.
\newblock Direct preference optimization: Your language model is secretly a reward model.
\newblock \emph{arXiv preprint arXiv:2305.18290}.

\bibitem[{Schulman et~al.(2017)Schulman, Wolski, Dhariwal, Radford, and Klimov}]{schulman2017proximal}
John Schulman, Filip Wolski, Prafulla Dhariwal, Alec Radford, and Oleg Klimov. 2017.
\newblock Proximal policy optimization algorithms.
\newblock \emph{arXiv preprint arXiv:1707.06347}.

\bibitem[{Teknium(2023)}]{openhermes}
Teknium. 2023.
\newblock teknium/openhermes-2.5-mistral-7b.
\newblock \url{https://huggingface.co/teknium/OpenHermes-2.5-Mistral-7B}.

\bibitem[{Tunstall et~al.(2023)Tunstall, Beeching, Lambert, Rajani, Rasul, Belkada, Huang, von Werra, Fourrier, Habib et~al.}]{tunstall2023zephyr}
Lewis Tunstall, Edward Beeching, Nathan Lambert, Nazneen Rajani, Kashif Rasul, Younes Belkada, Shengyi Huang, Leandro von Werra, Cl{\'e}mentine Fourrier, Nathan Habib, et~al. 2023.
\newblock Zephyr: Direct distillation of lm alignment.
\newblock \emph{arXiv preprint arXiv:2310.16944}.

\bibitem[{Upstage(2023)}]{solar70}
Upstage. 2023.
\newblock upstage/solar-0-70b-16bit.
\newblock \url{https://huggingface.co/upstage/SOLAR-0-70b-16bit}.

\bibitem[{Wei et~al.(2022)Wei, Tay, Bommasani, Raffel, Zoph, Borgeaud, Yogatama, Bosma, Zhou, Metzler et~al.}]{wei2022emergent}
Jason Wei, Yi~Tay, Rishi Bommasani, Colin Raffel, Barret Zoph, Sebastian Borgeaud, Dani Yogatama, Maarten Bosma, Denny Zhou, Donald Metzler, et~al. 2022.
\newblock Emergent abilities of large language models.
\newblock \emph{arXiv preprint arXiv:2206.07682}.

\bibitem[{Yuan et~al.(2024)Yuan, Pang, Cho, Sukhbaatar, Xu, and Weston}]{yuan2024self}
Weizhe Yuan, Richard~Yuanzhe Pang, Kyunghyun Cho, Sainbayar Sukhbaatar, Jing Xu, and Jason Weston. 2024.
\newblock Self-rewarding language models.
\newblock \emph{arXiv preprint arXiv:2401.10020}.

\bibitem[{Yuan et~al.(2023)Yuan, Yuan, Tan, Wang, Huang, and Huang}]{yuan2023rrhf}
Zheng Yuan, Hongyi Yuan, Chuanqi Tan, Wei Wang, Songfang Huang, and Fei Huang. 2023.
\newblock Rrhf: Rank responses to align language models with human feedback without tears.
\newblock \emph{arXiv preprint arXiv:2304.05302}.

\bibitem[{Zellers et~al.(2019)Zellers, Holtzman, Bisk, Farhadi, and Choi}]{zellers2019hellaswag}
Rowan Zellers, Ari Holtzman, Yonatan Bisk, Ali Farhadi, and Yejin Choi. 2019.
\newblock Hellaswag: Can a machine really finish your sentence?
\newblock In \emph{Proceedings of the 57th Annual Meeting of the Association for Computational Linguistics}, pages 4791--4800.

\bibitem[{Zheng et~al.(2023)Zheng, Chiang, Sheng, Zhuang, Wu, Zhuang, Lin, Li, Li, Xing et~al.}]{zheng2023judging}
Lianmin Zheng, Wei-Lin Chiang, Ying Sheng, Siyuan Zhuang, Zhanghao Wu, Yonghao Zhuang, Zi~Lin, Zhuohan Li, Dacheng Li, Eric Xing, et~al. 2023.
\newblock Judging llm-as-a-judge with mt-bench and chatbot arena.
\newblock \emph{arXiv preprint arXiv:2306.05685}.

\bibitem[{Ziegler et~al.(2019)Ziegler, Stiennon, Wu, Brown, Radford, Amodei, Christiano, and Irving}]{ziegler2019fine}
Daniel~M Ziegler, Nisan Stiennon, Jeffrey Wu, Tom~B Brown, Alec Radford, Dario Amodei, Paul Christiano, and Geoffrey Irving. 2019.
\newblock Fine-tuning language models from human preferences.
\newblock \emph{arXiv preprint arXiv:1909.08593}.

\end{thebibliography}

\end{document}